# Symbol-based entity marker highlighting for enhanced text mining in materials science with generative AI

Junhyeong Lee[1,2], Jong Min Yuk[3], and Chan-Woo Lee[1,4*]

1. Energy Storage Research Department, Korea Institute of Energy Research, Daejeon 34129, Republic of Korea

2. Department of Energy Engineering, Hanyang University, Seoul, 04763, Republic of Korea

3. Department of Materials Science and Engineering, Korea Advanced Institute of Science and Technology, Daejeon 34141, Republic of Korea

4. Energy AI & Computational Science Laboratory, Korea Institute of Energy Research, Daejeon 34129, Republic of Korea

*Corresponding author. E-mail: cwandtj@kier.re.kr

# Abstract

The construction of experimental datasets is essential for expanding the scope of data-driven scientific discovery. Recent advances in natural language processing (NLP) have facilitated automatic extraction of structured data from unstructured scientific literature. While existing approaches—multi-step and direct methods—offer valuable capabilities, they also come with limitations when applied independently. Here, we propose a novel hybrid text-mining framework that integrates the advantages of both methods to convert unstructured scientific text into structured data. Our approach first transforms raw text into entity-recognized text, and subsequently into structured form. Furthermore, beyond the overall data structuring framework, we also enhance entity recognition performance by introducing an entity marker—a simple yet effective technique that uses symbolic annotations to highlight target entities. Specifically, our entity marker-based hybrid approach not only consistently outperforms previous entity recognition approaches across three benchmark datasets (MatScholar, SOFC, and SOFC slot NER) but also improve the quality of final structured data—yielding up to a 58% improvement in entity-level F1 score and up to 83% improvement in relation-level F1 score compared to direct approach.

# Introduction

With the rapid advancement of computational technologies, large-scale databases with artificial intelligence (AI) have been increasingly employed to accelerate the scientific discovery and innovation[1–3]. Among various strategies, inverse design—an approach that begins with desired properties and works backward to identify or generate materials that exhibit them—has emerged as one of the most promising strategies[4–6], where AI models are developed to screen or generate innovative materials accordingly [7–9]. However, given the relative ease of generating large and consistent datasets through computational simulations, currently available datasets are often derived from the computational simulation [10–13]. This may, in part, explain the tendency of AI-based studies to make use of theoretical data, despite the significant potential of experimental data. To expand the scope of AI models, there is increasing demand for the construction of experimental datasets[14–17].

One of the effective approaches to building experimental datasets is to use natural language processing (NLP) algorithms to extract experimental knowledge from the scientific literature[18,19]. Through these algorithms, unstructured scientific texts can be converted into machine-readable, structured data. A common strategy for structuring data is to extract entities (e.g., *"cathode"* and *"LiMnPO4"*) and relations (e.g., *"cathode"* is application of *"LiMnPO4"*)[20–22]. Conventionally, structured data has been built through a multi-step pipeline (Fig. 1a); named entity recognition (NER)—where each relevant word is assigned a specific label (such as *"material"* or *"application"*)—is performed first, followed by relation extraction (RE)[23,24]. This approach has led to high entity recognition performance, particularly through the use of AI models in the NER step[25–29]. However, in the subsequent RE step, relatively straightforward techniques (e.g., rule-based approaches[30,31]) are often adopted, which may have limitations in capturing the complex relationships found in scientific texts. For instance, in the sentence *"nano platinum is used as a catalyst"*, such techniques may extract simple, pairwise relations such as *"platinum-catalyst"* or *"nano-platinum"*, rather than complex relationship such as *"nano-platinum-catalyst"*. To convey the intended meaning more effectively, the combination of all three elements is preferred.

More recently, to capture these complex relationships often required in scientific texts, researchers have turned to generative AI models—such as ChatGPT[32,33], LLaMA[34–36], Gemini[37,38], DeepSeek[39,40], and T5[41]—which demonstrate remarkable capabilities across various tasks: automating the generation and prediction of new novel materials[42] and extracting entities from scientific literature[43–45]. With these generative AI models, Dagdelen et al.[46] introduced direct approach (without intermediate steps), facilitating more effective extraction of complex relationships (Fig. 1b); they fed prompts into the model and directly transformed unstructured scientific text into structured data. Furthermore, building on this approach, Lee et al.[47] successfully analyzed the synthesis method of gold nanoparticles from scientific publications. However, despite the direct approach enables to capture complex relationships, compressing the NER and RE steps may sometimes cause mistakes in entity recognition.

To take advantage of both approaches, we propose a hybrid framework (Fig. 1c), consisting of two steps:

1. **The NER phase (as in the multi-step approach):** Each entity is recognized and highlighted within the text, producing an entity-recognized text that enhances entity recognition performance in the following step.
2. **Unstructured-to-structured phase (as in the direct approach):** Generative AI is leveraged to directly convert the entity-recognized text into structured data, thereby capturing complex relationships.

Beyond the overall data structuring framework, we also contribute at the component level—specifically, in the NER phase. We propose a simple yet effective technique: the use of entity markers. These markers are defined using entity symbols and are inserted into the unstructured raw text to explicitly highlight entity types (see entity-recognized text shown in Fig. 1c). In doing so, we combine the strengths of two existing NER paradigms: (1) encoder-only models, which extract multiple entity types simultaneously but learn solely from entity-labeled sentences[26,27,48], and (2) generative models with special markers, which learn entity descriptions via in-context learning but handle single entity type at a time[43,45,49,50]. By bridging these two approaches, our method delivers a dual goal: multi-type entity extraction and in-context learning of entity descriptions, thus improving NER efficiency and promoting a deeper understanding of entity meanings.

Ultimately, our two-level contributions—at both the component (i.e., within the NER phase) and overall level (i.e., across the full data structuring process)—works synergistically to enhance the following outcomes:

- **NER performance:** The entity marker-based method outperforms previous NER approaches across three benchmark datasets (MatScholar, SOFC, and SOFC slot NER dataset).
- **Structured data quality:** The hybrid framework leads up to 58% improvement in the F1 score for structured data entities, compared to direct approach. Furthermore, since relationships are built upon these entities, the relation F1 score also yields up to an 83% improvement.

The remainder of this paper is organized as follows. First, as a component-level contribution, we present an entity marker strategy. Next, as an overall-level contribution, we then introduce a streamlined workflow. Finally, we report the evaluation results for both two-level contributions, demonstrating the effectiveness of our proposed framework.

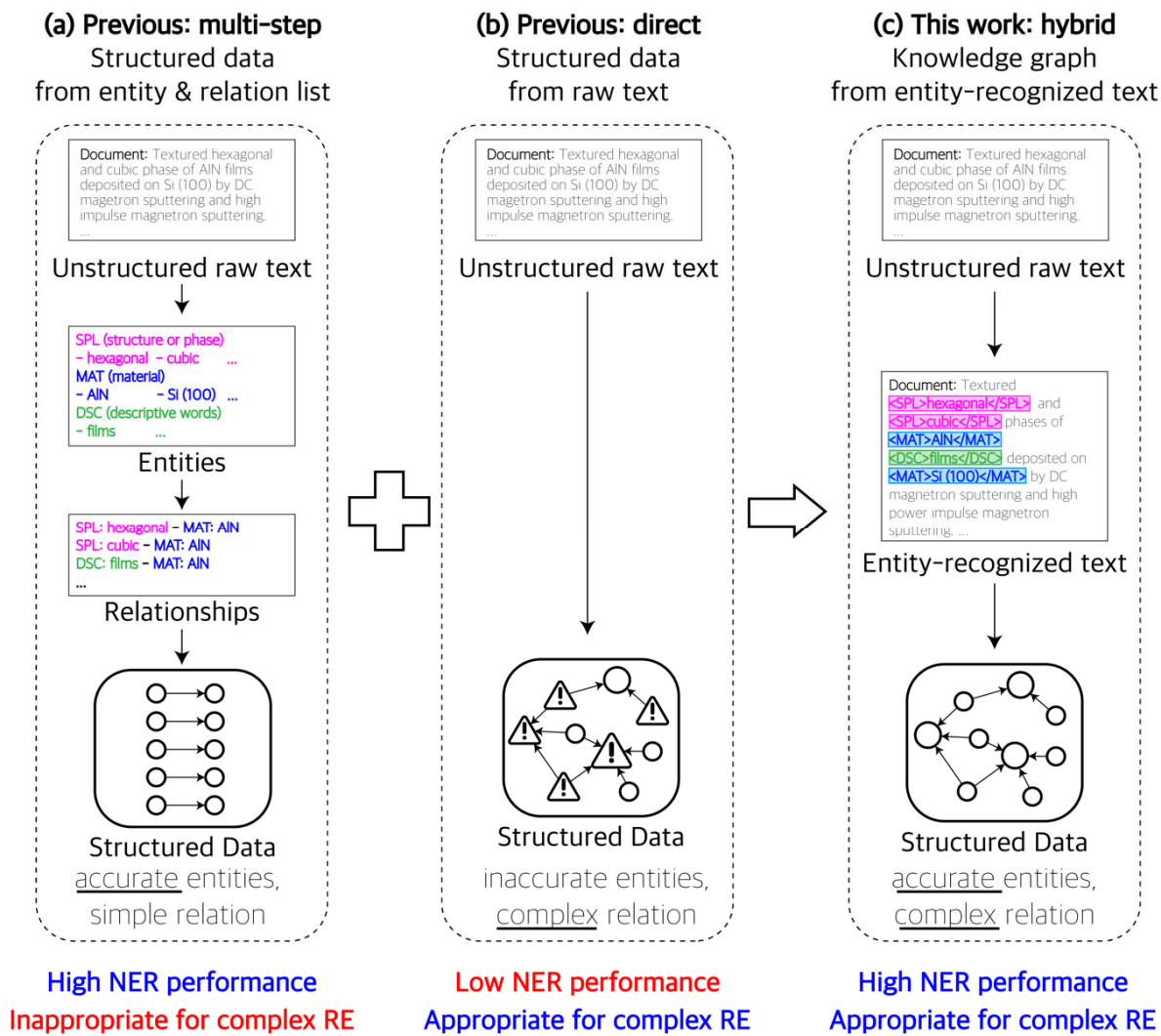

**Fig.1: Schematic comparison of the conventional methods of structured data construction with this work.**

**Results and discussion**

**Entity marker: enhancing named entity recognition (NER) performance and efficiency**

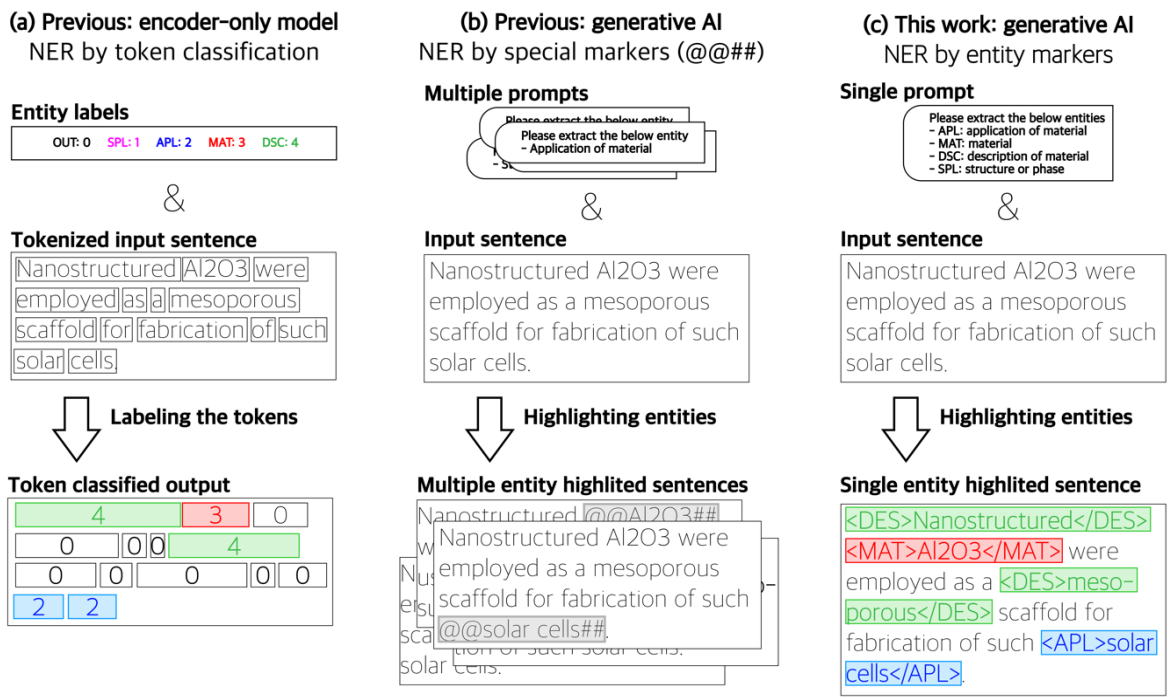

Fig.2: Schematic comparison of the conventional methods for NER with this work.

For the NER task, the encoder-only model approach[48] and the generative AI model approach utilizing special marker[49] show demonstrated performance in the previous studies. Nevertheless, despite their effectiveness, these methods have certain limitations. To overcome this, we proposed a new approach using entity markers (Fig. 2). Encoder-only models, such as BERT, are typically trained as token classification models (Fig. 2a)[49]. Through the model, each token—the smallest unit of text processed by the model—in the input sentences is analyzed and classified into a predefined entity category. However, since these categories are fixed before training, the model cannot recognize entities that were not included in the training dataset[43]. To introduce a new entity type, the model must be retrained; the dataset must be reconstructed with updated entities, where each token is manually labeled with the appropriate entity type.

On the other hand, an alternative strategy has been proposed and implemented for generative AI, since it exhibits different property compared to encoder-only models. In this strategy, the model is provided with entity definitions in the prompt and generates the entity highlighted sentence, where each entity is marked with special tokens—@@ before and ## after the entity (Fig. 2b). By employing in-context learning with entity definitions given in the prompt, the model is able to distinguish entities which are not present in the fine-tuning dataset; in-context learning allows the model to comprehend the description of entity introduced in the prompt, promoting a deeper understanding and thus eliminating the need for additional training whenever the desired data schema changes. However, since this approach relies on arbitrary markers, extracting multiple entities from a sentence requires multiple prompts, thereby increasing computational costs.

Addressing these limitations, our generative AI based NER method introduces entity marker using corresponding entity symbols—*<entity symbol>* before and *</entity symbol>* after (Fig. 2c). The adoption of symbol-based markers enables the model to detect multiple entities using single prompt while preserving in-context learning. In other words, it combines the advantage of previous techniques: (1) multi-type entity detection by processing only one prompt and (2) in-context learning to identify entities beyond fine-tuning dataset. Consequently, our approach not only enhances the flexibility of entity recognition but also improves processing efficiency by reducing the need for repeated prompts.

**Overview of workflow**

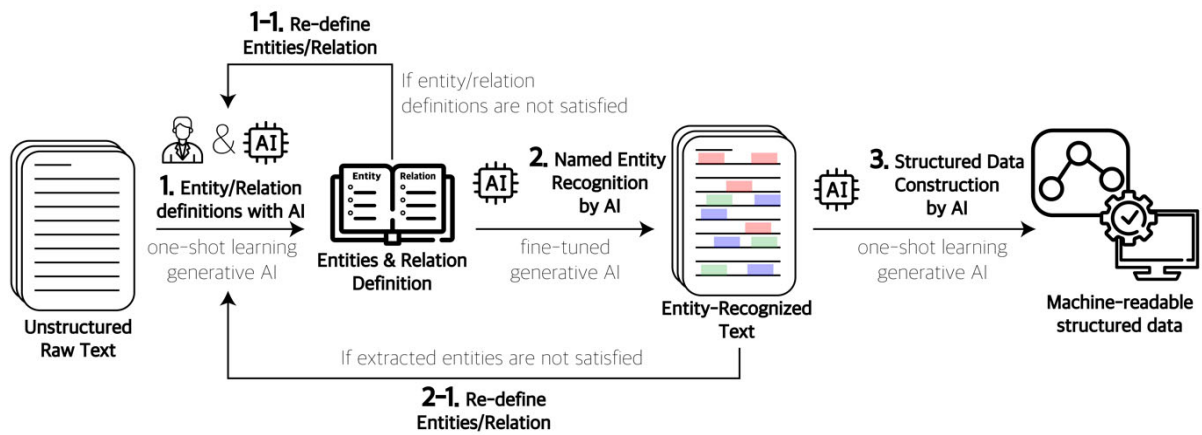

**Fig.3: Workflow for transforming unstructured raw text into machine-readable structured data via entity recognized text**

Overall workflow to construct machine-readable structured data consists of three key steps (Fig. 3). First a researcher interacts with a one-shot learning AI to define the entities and relationships to be extracted (step 1 and 1-1). During this process, entities are assigned their unique symbols with respective descriptions and relationships between these entities are explicitly specified (step 1). This step is iterative allowing the researcher to refine definitions through continuous feedback with AI (step 1-1). Next, a fine-tuned generative AI processes raw text and generates entity-recognized sentences where identified entities are highlighted with their entity markers (step 2). The researcher reviews these entity-recognized sentences and if necessary, the researcher returns to the definition step to refine the entity and relation definitions (step 2-1). Finally, once the researcher is satisfied with the entity-recognized sentences, the one-shot learning AI transforms them into a machine-readable structured data (step 3). Compared to step 2 (NER phase), constructing the fine-tuning datasets for step 3 is more challenging due to the diverse formats of the structured data depending on its purpose. Thus, adopting one-shot learning AI—where the model learns from a single example provided in a prompt without additional training—in step 3 eliminates the need to create fine-tuning datasets, a process that is time-consuming and requires expert manual effort.

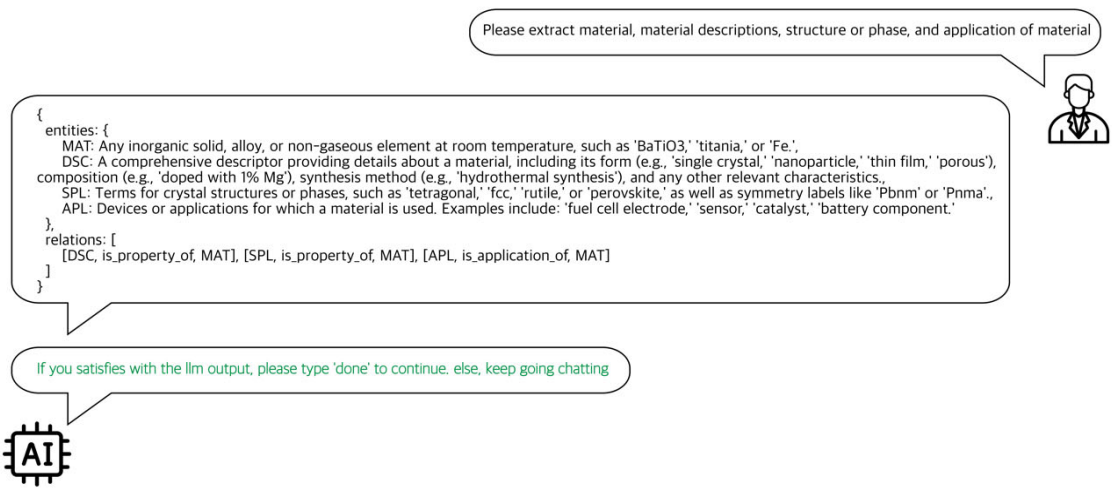

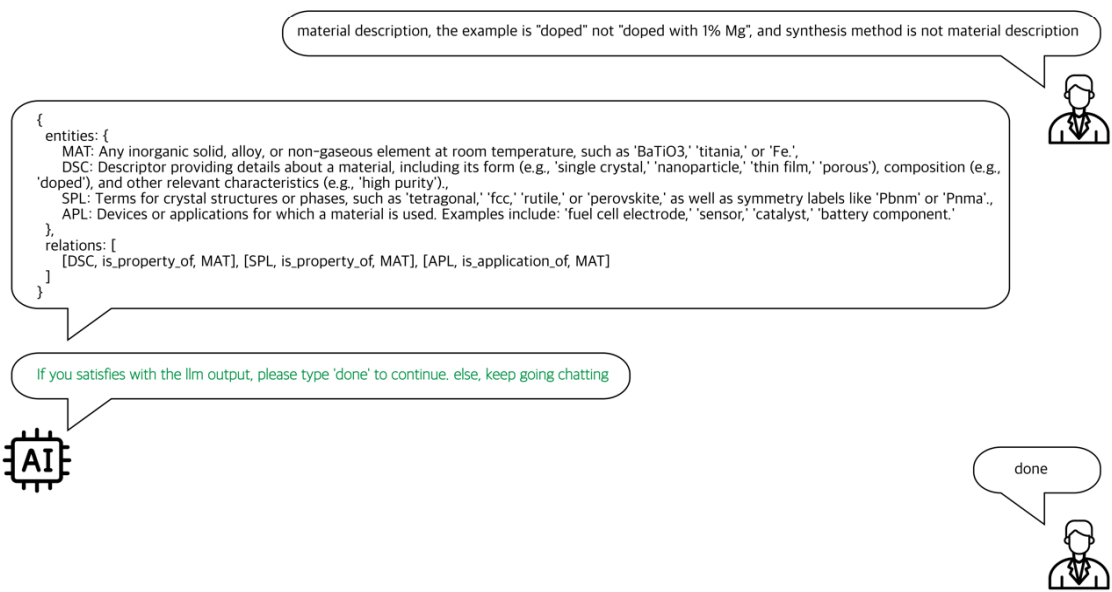

**Fig.4: Example for interaction between researcher and LLMs during entities and relations definition**

The interaction between the researcher and AI is the first step of the workflow, which involves: (1) generating preliminary entity and relation definitions based on the researcher's input, and

(2) refining these definitions through researcher-AI communication (Fig. 4). For example, when a prompt is provided with researcher's request to extract information on materials, their descriptions, structure or phases, and applications from research papers, an initial set of definition is generated (Fig. 4a). In this definition, the AI assigns entity symbols—MAT (material), DSC (material descriptions), SPL (structure or phase) and APL (application of materials)—along with a corresponding description, while simultaneously defines the relationships between these entities. Following the generation of the initial definition, additional researcher-AI interaction is carried out to refine the definition—for instance, by modifying the example for the material description entity to strengthen the accuracy and specificity (Fig. 4b). The detailed prompts along with explanations are provided in the supplementary information.

Figure 5 shows a schematic example of transforming unstructured raw text into entity-recognized text, followed by structured data construction. For entity-recognized text, a fine-tuned generative AI uses entity symbols and descriptions provided in the prompt to identify and highlight entities within the text (Fig. 2c). After the generation of entity-recognized sentences, the researcher reviews the output to ensure accuracy. If the AI misidentifies certain entities, the entity descriptions could be refined to enhance the AI's in-context learning performance. In such case, the researcher may revisit the interaction step (Fig. 4b) to optimize entity recognition. Finally, once the entity-recognized sentences meet the desired accuracy, they are used to construct the structured data.

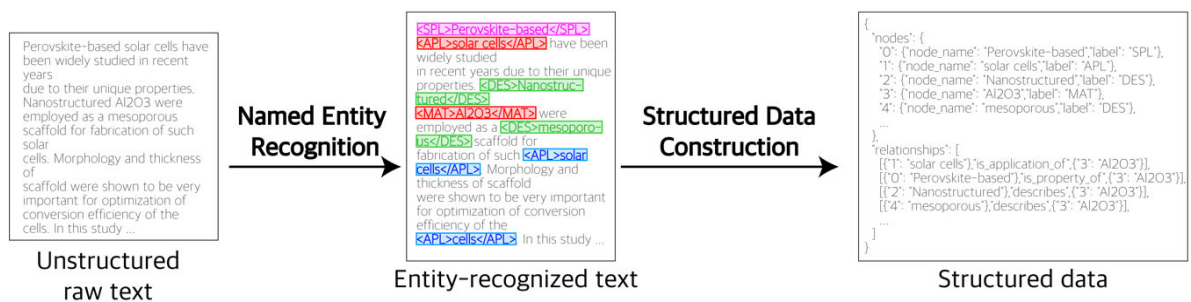

**Fig.5: Example for structured data construction from unstructured raw text via entity recognized text**

**Comparison of NER performance**

To assess the effectiveness of our proposed entity marker approach in the NER phase (Fig. 2c), we fine-tuned and evaluated the AI models on several datasets (see Methods section for fine-tuning details and Supplementary information for datasets details). In the NER task, reducing both wrong predictions and valid entity omissions is important. Therefore, we reported precision (which measures how many predicted entities are correct), recall (which measures how many entities are identified) and F1 score (the harmonic means of precision and recall) across three approaches: the encoder-only model approach, the special marker approach and the entity marker approach (Fig. 6). In addition, we adopted exact match criteria to evaluate the training performance of each model. Further details of the evaluation process are provided in Method section.

Encoder-only models frequently misclassify unrelated tokens as entities, resulting in the lowest precision (Fig. 6a). This occurs because they rely solely on entity patterns observed in entity-labeled sentences rather than comprehending the semantic meaning of entities. In contrast, special marker approach tends to omit the valid entities, leading to the lowest recall (Fig. 6b). Because the special marker approach requires a separate prompt for each entity type, fine-tuning sentences contain fewer highlighted entities, making the model more conservative in entity detection (see supplementary information). In other words, the model learns to highlight the entities more cautiously. However, the entity marker approach overcomes the limitations of the encoder-only and special marker methods—namely, misclassification and valid entities omission, respectively—by balancing precision and recall, thereby achieving the highest F1 score (Fig. 6c). Compared to encoder-only models, but similar to special marker approach, it leverages a deep comprehension of entity meaning through in-context learning. Furthermore, in contrast to the special marker approach, while being similar to encoder-only models, it allows the model to classify multi-type entities using a single prompt, resulting in more highlighted entities within fine-tuning sentences and preventing model from becoming conservative. This synergy yields the highest F1 score across all datasets while also minimizing error bars, which indicates the training process is more stable.

Moreover, two key factors of our entity marker method— (1) in-context learning and (2) multi-type entity extraction—demonstrate substantial performance gains on the challenging SOFC Slot NER dataset (see supplementary discussion for details on dataset difficulty). First, in-context learning allows the model to develop a deeper understanding of entities beyond their patterns found in labeled sentences. This leads to improved performance of generative based approaches—such as special marker or entity marker methods—compared to encoder-only model approaches. Second, multi-type entity extraction makes the model learn various types of entities simultaneously. This is especially beneficial in scientific domains like materials research, where jointly learning oppositional entities (e.g., anode vs cathode) is more effective than learning them separately. Consequently, the entity marker method achieved further improvements on the SOFC Slot NER dataset, which contains several such oppositional entity pairs.

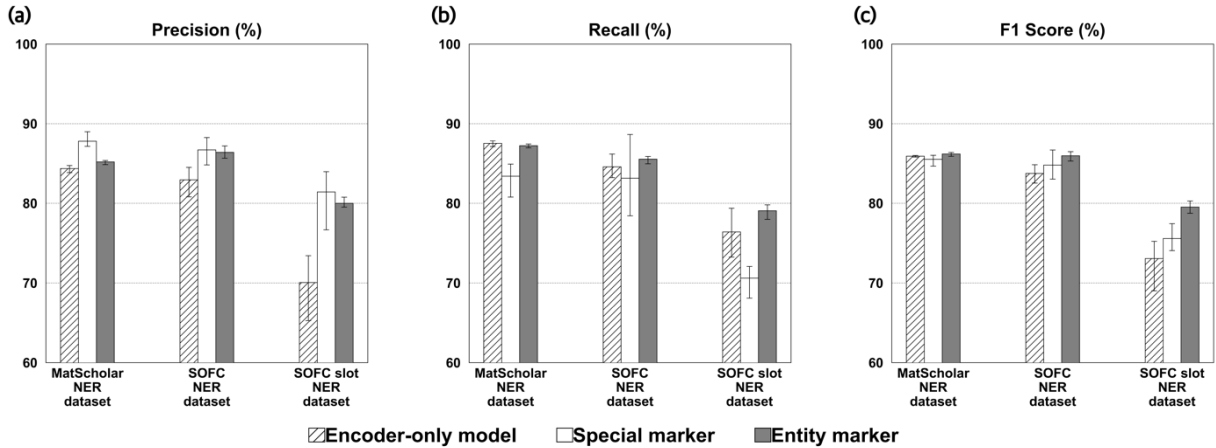

Fig.6: Exact match (a) precision, (b) recall, and (c) F1 score on various test datasets (MatScholar, SOFC, and SOFC-Slot).

To evaluate the efficiency of different NER approaches, the relative inference time—defined as the inference time divided by the inference time of the entity marker approach—was measured (Fig. 7.) The entity marker approach significantly reduces inference time compared to the special marker approach, as it enables the simultaneous extraction of multiple entities within a single prompt whereas the special marker approach requires multiple prompts, with the number of prompts increasing proportionally to the number of entit. As a result, the inference time difference between the special marker approach and entity marker approach increases with dataset complexity, following the order of SOFC (four entities), MatScholar (seven entities) and SOFC Slot (18 entities) NER datasets. On the other hand, the encoder-only model operates over 100 times faster than generative AI approach, primarily due to the fundamental difference in their inference mechanism. While encoder-only models process the entire input sentence in a single step, generative AI models generate tokens sequentially, requiring multiple inference steps for completion. Despite its significantly fast inference, the efficiency of the encoder-only model is constrained by the need of additional training whenever the desired data schema changes. In contrast, generative AI models, while slower, benefit from in-context learning to handle to new data schemas without additional training, thereby optimizing their adaptability and ultimately their overall efficiency in the NER task.

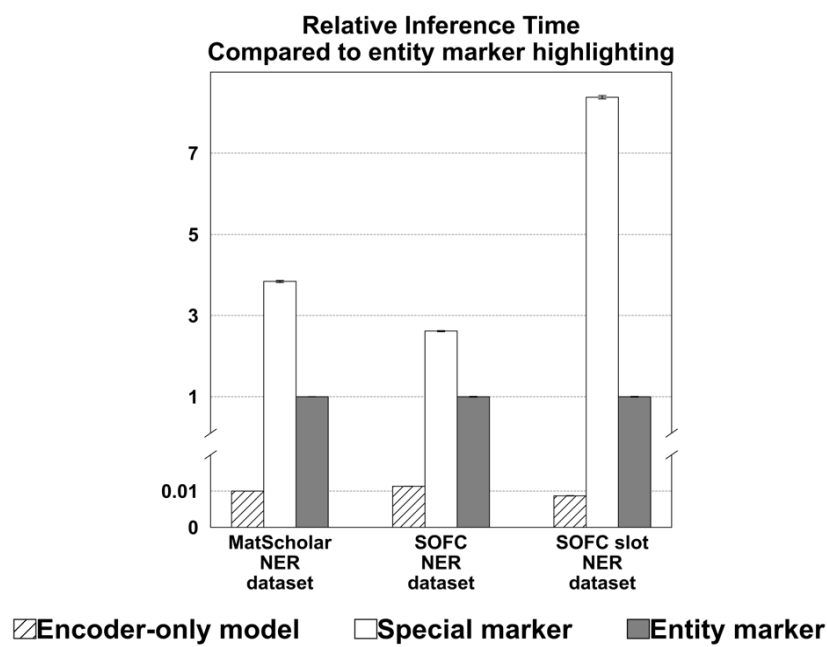

**Fig. 7: Relative inference time on the test set for MatScholar, SOFC, and SOFC-Slot datasets.**

**Comparison of structured data construction performance**

For the evaluation of structured data construction, knowledge graphs are chosen as the representative format due to their well-defined structure, which enables precise assessment of AI's entity recognition and relation identification capabilities (see Methods section). To construct the knowledge graph including complex relationships, two approaches—direct (Fig. 1b) and hybrid (Fig. 1c)—are applied. In addition, the evaluation process adopts manual annotation to capture nuances that an exact match approach might overlook, ensuring a more comprehensive assessment of structured data quality, as the detailed explanation in supplementary information.

Entity-recognized text helps AI distinguish entities from context, thereby minimizing error propagation—where mistakes in entity (or node) recognition affect relation extraction. This is why the hybrid approach yields a higher F1 scores compared to the direct approach (Fig. 8), even though both direct and hybrid approaches follow the same process of transforming the unstructured text. Specifically, highlighting entities simplifies distinguishing them, which in turn reduces the overall complexity of the conversion process and leads a higher node F1 score for the hybrid approach (Fig. 8a). Given the inherent complexity of the task—which involves not only the identification of entities but also the analysis of their relationships—such simplification is important for generating high-quality structured data. Furthermore, by suppressing error propagation, the hybrid approach achieves higher F1 score on relationships (Fig. 8b). Because the relationships are established based on the detected entities, any errors in node propagate to relationships. As a result, the F1 score for relationships tend to be lower than that of nodes. This pattern of error propagation is also reflected in specific relationships; for example, while the *SPL is property of MAT* relationship achieves the highest F1 score due to high recognition performance of both nodes, other relationships show lower scores due to underperformance in at least one associated node. To improve the F1 score on relationships, strengthening AI's node recognition ability is crucial, as it regulates the error propagation. In summary, implementing entity highlighting before building the knowledge graph significantly facilitates a higher F1 score for nodes, which in turn enhances the relationship identification performance.

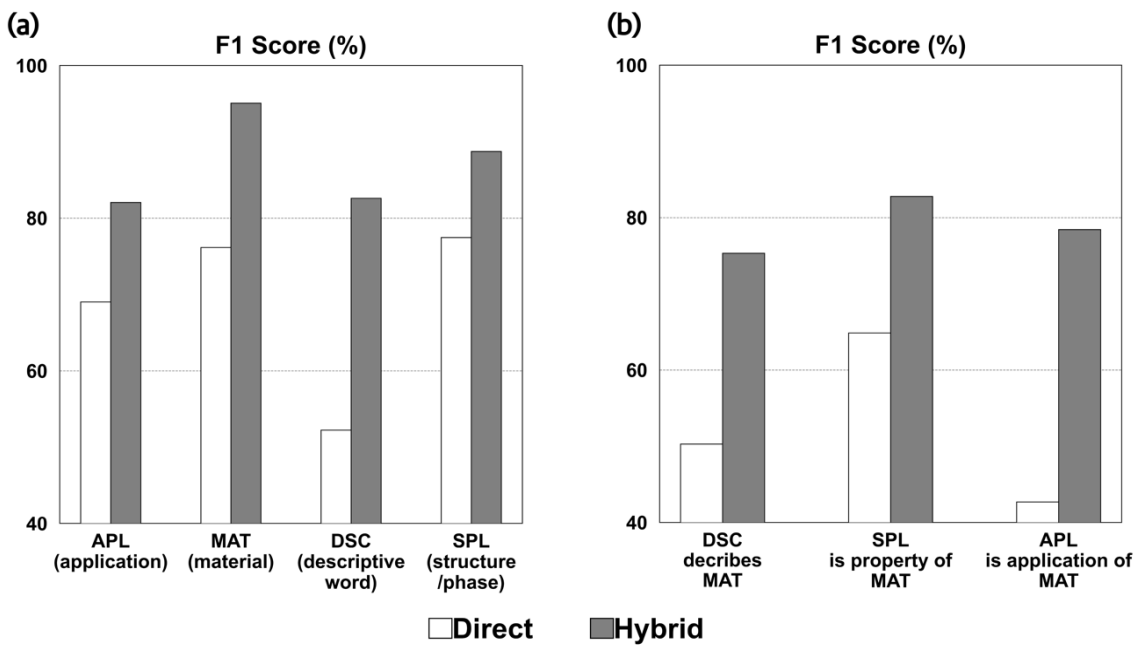

**Fig. 8: Manually annotated F1 scores for their nodes and relations**

## Conclusion

In this study, we proposed novel text mining approach, achieving two-level contributions. First, at the component-level (i.e., within the NER phase), we introduced a simple yet effective concept, entity marker, which is effectively addressing two major challenges in NER phase: misclassification and valid entity omission. By leveraging (1) in-context learning—which allows the model to learn entity descriptions—and (2) multi-type entity extraction—which maximizes the number of highlighted entities within fine-tuning sentences—the model can develop a deeper semantic understanding of entities while also avoiding conservative predictions. As a result, the entity marker approach balances precision and recall, delivering the highest F1 scores across all evaluated datasets. In addition, the synergy of in-context learning and multi-type entity extraction further enhances the model's ability to recognize oppositional entities, which is particularly advantageous in scientific domains such as materials research. This led to notable performance gains on the challenging SOFC Slot NER dataset, which includes such oppositional entities.

Next, we integrated existing data structuring techniques into a hybrid framework. Similar to the approach by Dagdelen et al.[46], a generative AI model was employed to directly convert unstructured text into a structured format, supporting the extraction of complex relationships often found in scientific literature. However, prior to this conversion, our approach applied NER to transform raw text into entity recognized text, with its unstructured format preserved but entities being highlighted. As a result, the conversion process was simplified, which in turn effectively reduced error propagation. Specifically, the presence of highlighted entities in entity-recognized text allowed the AI model to more easily identify structured data entities, leading to an improved entity-level F1 score. Subsequently, the relation-level F1 score—being dependent on accurate entity recognition—was also enhanced.

## Methods

**Fine tuning language models on NER**

To perform the NER task, we fine-tuned both encoder-only language models—BERT-uncased[48], MatBERT-uncased[27], MatBERT-cased[27], and MatSciBERT[26]—and a generative AI model, LLaMA[34–36]. To optimize NER performance of encoder-only models, we added a CRF layer to the output layer[51–53] of each model using PyTorch-CRF library[54]. While each BERT model was fine-tuned independently on one of the NER datasets, single LLaMA model was fine-tuned on multiple NER datasets.

The following materials science-specific and general-purpose NER datasets were used to train the language models for the NER task:

1. **MatScholar NER dataset**[25]**:** This dataset focuses on material science literature and includes seven entity types: material (MAT), structure of phase (SPL), material descriptions (DSC), application of material (APL), property of material (PRO), synthesis method (SMT), and characterization method (CMT). It contains 4401, 511, and 546 sentences in training, validation and test sets, respectively.

2. **SOFC NER dataset**[55]**:** This dataset is derived from materials text related to solid oxide fuel cells (SOFCs). It includes four entity types: EXPERIMENT, VALUE, MATERIAL and DEVICE. The training, validation and test sets consist of 568, 135, and 173 sentences, respectively.

3. **SOFC Slot NER dataset**[55]**:** This dataset contains same sentences with SOFC NER dataset, but with a more fine-grained entity labeling scheme. It includes 18 entity types: anode material, cathode material, conductivity, current density, degradation rate, device, electrolyte material, experimental evoking word, fuel used, interlayer material, open circuit voltage, power density, resistance, support material, thickness, time of operation, voltage, and working temperature. These labels reflect more specific and complex information relative to SOFC NER dataset.

4. **CoNLL 2003 dataset**[56]**:** This dataset involves general-purpose sentences which is composed of four entities: person (PER), organization (ORG), location (LOC) and miscellaneous entity (MISC). To enhance generative AI model's ability to comprehend a broader range of entity types, this dataset was included exclusively in its training; encoder-only models were not trained on this dataset.

For simplifying the preprocessing process, the MatScholar and SOFC datasets were formatted and stored using HuggingFace datasets library [57–61]; The existing CoNLL-2003 dataset was used without modification[62].

To train BERT as a token classification model, entities were encoded using the BIO tagging scheme to account for multi-token entities. In this scheme, each token in a sentence is assigned one of the following labels:

1. **B-X**, which indicates the beginning of an entity of type X
2. **I-X**, which denotes a token inside an entity of type X
3. **O**, which is used for tokens that do not belong to any named entity

For example, in the tokenized sentence [*Nanostructured*, *Al$_2$O$_3$*, …, *such*, *solar*, *cells*], the corresponding labels are [*B-DSC*, *B-MAT*, …, *O*, *B-APL*, *I-APL*] where *"Nanostructured"* is labeled as a material description (DSC), *"Al$_2$O$_3$"* as a material (MAT), and *"solar cells"* as an application of material (APL).

LLaMA was trained to comprehend descriptions of entity types provided in prompts and to identify and highlight the corresponding entities within input text. To improve the flexibility and generality of NER task, various descriptions were generated. These descriptions convey the same semantic meaning of each entity type but differ in wording. For example, two representative descriptions of the entity structure or phase (SPL) are:

- Terms for crystal structures or phases, such as 'tetragonal,' 'fcc,' 'rutile,' or 'perovskite,' as well as symmetry labels like 'Pbnm' or 'Pnma.'
- Names of crystal structures or phases, including 'tetragonal,' 'fcc,' 'rutile,' and 'perovskite,' or symmetry classifications like 'Pbnm' and 'Pnma.'

Approximately 10 descriptions were generated per entity type using ChatGPT[33] and subsequently refined through manual curation (see supplementary information for the full list of descriptions). For model training, different labeling strategies were used. In the special marker approach, the same sentence was duplicated for each entity type, and entities of that type were highlighted within the sentence using special marker (Fig. 2b). In contrast, the entity marker approach did not involve sentence duplication; instead, all entities of multiple entity types were simultaneously highlighted within the sentence using entity marker (Fig. 2c). Prompts were constructed by inserting the selected entity descriptions and input text into predefined template (see supplementary information), while the completions consisted of entity highlighted sentence using either special markers or entity markers.

**Structured data construction of generative AI model**

To transform unstructured text into structured data, we adopted a pretrained generative AI model—the Q4_K_M pre-quantized 70-billion parameter version of LLaMA-3.3[63]. From the dataset constructed by Dagdelen et al.[46], we selected 49 abstracts and manually built knowledge graphs in JSON format. In their dataset, structured data was prepared through a *"human-in-the-loop"* process, where intermediate model suggestions were iteratively corrected and used to retrain the model.

In the knowledge graphs, entities are stored in nodes and their relationships in edges. When an entity is referred to multiple times with different expressions, the most descriptive term is chosen as the node name, and alternative expressions are recorded under a *"co_references"* field. Although nodes are primarily distinguished by their names, we introduced an additional *"related_entities"* field to accommodate cases where separation based solely on naming is ambiguous. For example, when the entity "LiMnPO$_4$" (also referred to as LMFP) appears in different synthesis contexts such as *solvothermal* or *co-precipitation*, the corresponding nodes are represented as:

- *{"node_name": 'LiMn0.9Fe0.1PO4', "co_references": ["LMFP"], "related_entities": [{"SYN": "solvothermal"}]}*
- *{"node_name": 'LiMn0.9Fe0.1PO4', "co-references": ["LMFP"], "related_entities": [{"SYN": "co-precipitation"}]}*

Further details regarding the prompt design are provided in the supplementary information.

**Performance evaluation**

To assess misclassification and valid data omission of the models, precision, recall and F1 score were reported which are calculated as:

$$\text{precision} = \frac{\text{TP}}{\text{TP} + \text{FP}}, \text{recall} = \frac{\text{TP}}{\text{TP} + \text{FN}}, \text{F1} = \frac{2(\text{precision} \cdot \text{recall})}{\text{precision} + \text{recall}}$$

where TP, FP and FN represent true positive, false positive, and false negative, respectively. Precision represents the proportion of correct outputs among all predictions, while recall reflects the proportion of correct outputs among all ground-truth instances. Specifically, a model with high precision produces highly relevant results with high confidence, even at the risk of missing some valid ones, whereas a model with high recall identifies as many instances as possible, even if some incorrect ones are included. Since precision and recall often exhibit a trade-off, the F1 score is used to provide a harmonic mean that balances the two metrics.

In the NER task, entities were extracted differently depending on the model type. For BERT-based models, entities were extracted using the BIO tagging scheme, where entities are identified as single tokens labeled with a B-X tag or as contiguous token sequences beginning with a B-X tag followed by one or more I-X tags. In contrast, for generative AI models, entities were extracted using regular expressions applied to entity-recognized text in which entity spans were highlighted with predefined markers. The predictions are evaluated using exact match criteria: a prediction was counted as a true positive if it exactly matched a ground-truth entity, and as a false positive otherwise. The predictions are evaluated using exact match criteria: a prediction was counted as a true positive if it exactly matched a ground-truth entity, and as a false positive otherwise.

Since entities and relations are stored separately as nodes and edges in the knowledge graph, we separately evaluate entity recognition and relation identification capabilities in context of structured data construction. A manual annotation process was adopted for this evaluation. A predicted node was counted as a true positive if its contextual meaning—based on *node_name* or *co_references*—is equivalent to a predefined node, and as a false positive if not. Similarly, a predicted relation was treated as a true positive if it correctly connected nodes present in the knowledge graph; other predictions were false positives.

In both evaluation settings—exact match and manual annotation—unmatched ground-truth entities, nodes, or relations were regarded as false negatives.

**Fine tuning hyperparameters**

The encoder-only models were fine-tuned using the HuggingFace Transformers library[57]. A linear learning rate scheduler with 100 warmup steps and a weight decay of 0.01 was used. To account for the random initialization of non-BERT layers, different learning rates were applied to the BERT and non-BERT components: the peak learning rate was set to 5e-4 for the BERT part and 3e-4 for the non-BERT part, using AdamW optimizer[26]. The models were trained for 30, 80, and 100 epochs with batch sizes of 128, 16, and 16 for MatScholar, SOFC, and SOFC Slot NER datasets, respectively. F1 score was selected as the representative validation metric.

Early stopping was applied if the F1 score did not improve over 10 epochs, and the checkpoint with the highest metric on the validation set was saved.

The LLaMA model was fine-tuned using Unsloth library[64]. Similar to the encoder-only model fine-tuning, the optimization strategy included a linear learning rate scheduler with 100 warmup steps and weight decay of 0.01. A pre-quantized 4-bit[65], 3 billion instruct version of LLaMA-3.2[66] was used and loaded with bfloat16 datatype. Parameter-efficient fine-tuning (PEFT) was employed via rank-stabilized low-rank adaptation (rsLoRA)[67,68], with LoRA rank $r=32$, scaling factor $\alpha=32$, and dropout rate of 0. The optimizer used was the paged 8-bit AdamW optimizer[69] applied with bfloat16 mixed precision. Fine-tuning was conducted for 2000 and 7500 training steps with a batch size of 32 for the special marker and entity marker approaches, respectively. Validation was performed every 500 steps, based on LLaMA loss. Early stopping was triggered if no improvement is observed for three consecutive validations, and the checkpoint with the lowest validation loss was saved. To prevent overfitting to non-highlighted completions, 50% of such examples were removed from the training dataset. Additionally, to optimize training efficiency, the maximum input sequence length was limited to 2048 tokens.

We trained each model type—BERT-based models and LLaMA—three times under each condition and reported the mean and deviation of their NER performance.

**Generation hyperparameters**

To evaluate the training performance of our fine-tuned generative AI models on NER, we utilized Unsloth library[64] to generate entity-highlighted sentences. During this process generation was performed with a temperature of 0.1 and top-p value of 0.9 (Table 1).

For the structured data construction task in our hybrid approach, the fine-tuned model were first converted into GGUF format using bfloat 16 datatype, and subsequently quantized to the Q4_K_M level via the llama.cpp framework[70].

For the conversion of abstracts into entity-highlighted sentences or knowledge graph, prompt chaining was implemented with LangChain and LangGraph library[71]. Text completions were then generated through Ollama framework[72] integrated with LangChain. For both tasks—entity highlighting and transformation of unstructured text—were performed with a temperature was set to 0 and top-p value to 0.9 (Table 1).

|  |  |  | temperature | top-p |
|---|---|---|---|---|
| **NER training** | **special marker** |  | 0.1 | 0.9 |
|  | **entity marker** |  | 0.1 | 0.9 |
| **Converting abstract** | **direct** | Knowledge graph construction | 0 | 0.9 |
|  | **hybrid** | Entity highlighting | 0 | 0.9 |
|  |  | Knowledge graph construction | 0 | 0.9 |

**Table 1. temperature and top-p values used in the generation tasks**